\documentclass[letterpaper]{article} % DO NOT CHANGE THIS
\usepackage{aaai23}  % DO NOT CHANGE THIS
\usepackage{times}  % DO NOT CHANGE THIS
\usepackage{helvet}  % DO NOT CHANGE THIS
\usepackage{courier}  % DO NOT CHANGE THIS
\usepackage[hyphens]{url}  % DO NOT CHANGE THIS
\usepackage{graphicx} % DO NOT CHANGE THIS
\usepackage{natbib}  % DO NOT CHANGE THIS AND DO NOT ADD ANY OPTIONS TO IT
\usepackage{caption} % DO NOT CHANGE THIS AND DO NOT ADD ANY OPTIONS TO IT
\usepackage{algorithm}
\usepackage{algpseudocode}
\usepackage{amsmath,amsfonts,bm}

\def\eqref#1{equation~\ref{#1}}
\def\1{\bm{1}}

\def\vt{{\bm{t}}}

\DeclareMathAlphabet{\mathsfit}{\encodingdefault}{\sfdefault}{m}{sl}
\SetMathAlphabet{\mathsfit}{bold}{\encodingdefault}{\sfdefault}{bx}{n}

\def\sC{{\mathbb{C}}}
\def\sD{{\mathbb{D}}}
\def\sO{{\mathbb{O}}}

\usepackage{color}

\usepackage{comment}
\usepackage{booktabs}
\usepackage{multirow}

\definecolor{Green}{RGB}{30,148,55}

\usepackage{newfloat}
\usepackage{listings}
\DeclareCaptionStyle{ruled}{labelfont=normalfont,labelsep=colon,strut=off} % DO NOT CHANGE THIS
\floatstyle{ruled}
\newfloat{listing}{tb}{lst}{}
\floatname{listing}{Listing}
\title{Emergent Quantized Communication}
\author {
    Boaz Carmeli, %\textsuperscript{\rm 1}
    Ron Meir, %\textsuperscript{\rm 1}
    Yonatan Belinkov%\textsuperscript{\rm 1}
    \thanks{Supported by the Viterbi Fellowship in the Center for Computer Engineering at the Technion.}
}
\affiliations {
    Technion -- Israel Institute of Technology\\
    boaz.carmeli@campus.technion.ac.il, rmeir@ee.technion.ac.il, belinkov@technion.ac.il
}
\begin{document}

\maketitle

\begin{abstract}
The field of emergent communication aims to understand the characteristics of communication as it emerges from artificial agents solving tasks that require information exchange.
Communication with discrete messages is considered a desired characteristic, for both scientific and applied reasons.
However,  training a multi-agent system with discrete communication is not straightforward, requiring either reinforcement learning algorithms or relaxing the discreteness requirement via a continuous approximation such as the Gumbel-softmax.
Both these solutions result in poor performance compared to fully continuous communication.
In this work, we propose an alternative approach to achieve discrete communication -- quantization of communicated messages.
Using message quantization allows us to train the model end-to-end, achieving superior performance in multiple setups.
Moreover, quantization is a natural framework that runs the gamut from continuous to discrete communication.
Thus,  it sets the ground for a broader view of multi-agent communication in the deep learning era.

\end{abstract}
\section{Introduction}

A key aspect in emergent communication systems is the channel by which agents communicate when trying to accomplish a common task. Prior work has recognized the importance of communicating over a discrete channel \cite{havrylov2017emergence_20,lazaridou2020emergent_1, vanneste2022analysis_103}. From a scientific point of view, investigating the characteristics of communication that emerges among artificial agents may contribute to our understanding of human language evolution. And from a practical point of view, discrete communication is required for natural human--machine interfaces. Thus, a large body of work has been concerned with enabling discrete communication in artificial multi-agent systems \cite[][\it inter alia]{foerster2016learning_10,havrylov2017emergence_20}. 
However, the discretization requirement poses a significant challenge to \emph{neural} multi-agent systems, which are typically trained with gradient-based optimization. Two main approaches have been proposed in the literature for overcoming this challenge, namely using reinforcement learning (RL) algorithms~\cite{williams1992simple_a4,lazaridou2016multi_14} or relaxing the discrete communication with continuous approximations such as the Gumbel-softmax \cite{jang2016categorical_a2, havrylov2017emergence_20}. The RL approach maintains discreteness, but systems optimized with the Gumbel-softmax typically perform better in this setting. However, Gumbel-softmax training is effectively done with continuous communication. Both discretization approaches perform far worse than a system with fully continuous communication. In short, the more discrete the channel, the worse the system's performance. 

\begin{figure}[t]
\centering
\includegraphics[width=\linewidth]{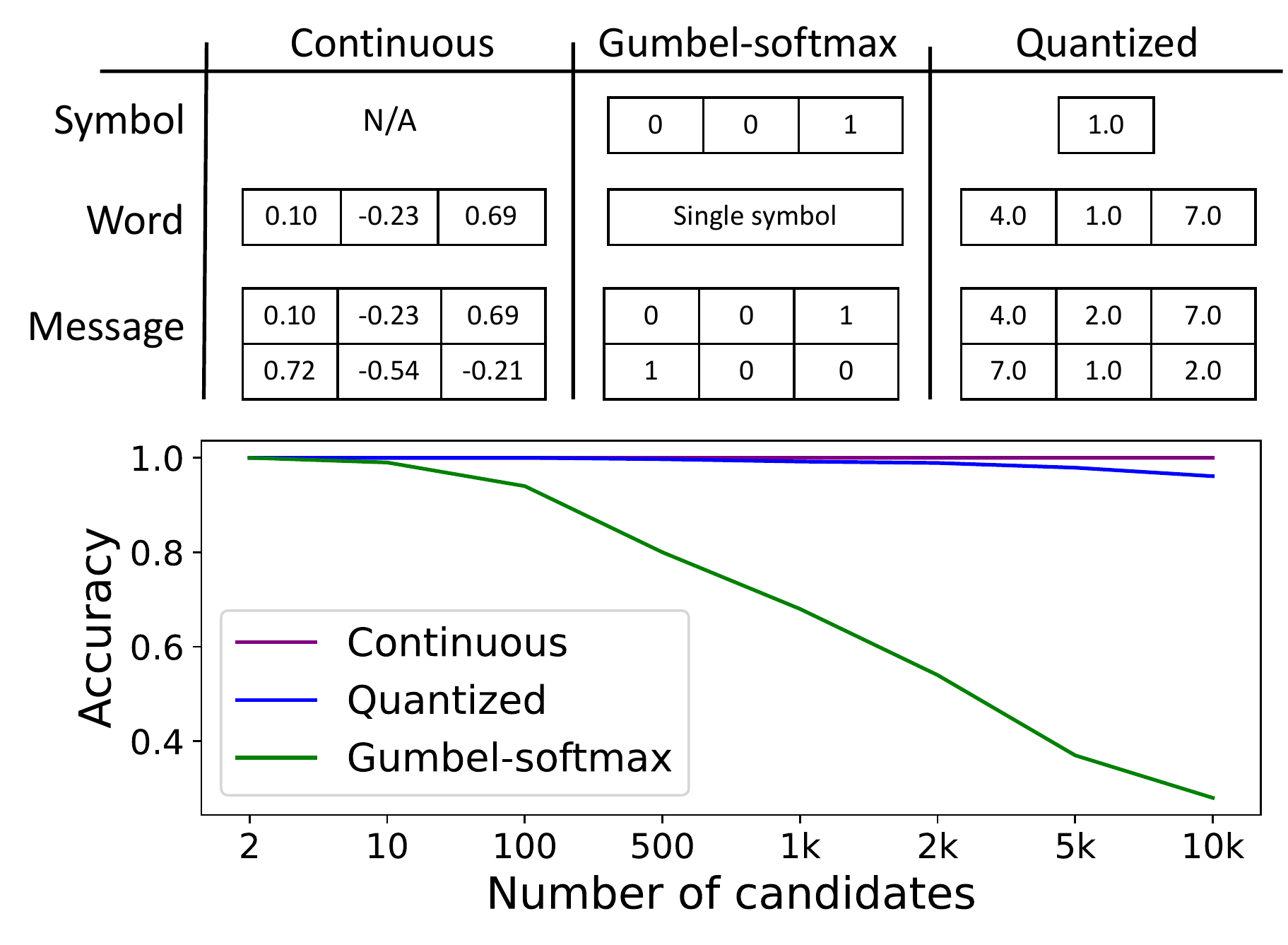}
\caption{\textbf{Top}: Symbol, word, and message elements for continuous, Gumbel-softmax, and quantized communication modes.
\textbf{Bottom}: Accuracy (Y-axis) achieved by the three communication modes vs.\! number of candidates (X-axis), in the Object game.
Continuous communication leads to good performance on the end task but does not use symbols. Gumbel-softmax sends one word per symbol, but requires  a recurrent channel and does not work well in practice. Quantized communication enables discrete and successful communication. Detailed channel parameters are provided in section \ref{sec:results}. 
}
\label{fig:channel}
\end{figure}

In this work, we propose a new framework for discrete communication in multi-agent systems, based on quantization (Figure~\ref{fig:channel}, top).
Drawing inspiration from work on efficient neural network quantization during training and inference  \cite{banner2018scalable_a9, wang2018training_a10, choi2018bridging_a16}),
we quantize the message delivered between the agents.
We investigate two learning setups: First, training is done with continuous communication, while inference is discretized by quantization, similar to the common scenario when using continuous approximations like Gumbel-softmax.
Second, we investigate the effects of quantizing the messages during both training and inference.

We experimentally validate our approach in multiple scenarios.
We consider three different games that fall into the well-known design of referential games, where a sender transmits information about a target object, which a receiver needs to identify \cite{lewis2008convention_a3, lazaridou2016multi_14, choi2018compositional_38, guo2019emergence_52}. Our objects include synthetic discrete objects, images, and texts. We also experiment with a variant, which we call the classification game, where the receiver needs to identify the class to which the object belongs. In all cases, we find our quantized communication to outperform the standard approach using Gumbel-softmax by a large margin, often even approaching the performance with fully continuous communication (Figure~\ref{fig:channel}, bottom). 

Finally, we investigate the quantized communication by varying the granularity of quantization. This allows us to cover a much wider range of discreteness levels than has previously been possible. We analyze which aspects of the communication channel are most important for accomplishing the agents' task and how they affect the resulting language. 
We find that quantization, even an extreme one, works surprisingly well given a long enough message.
Evidently, quantized communication with a binary alphabet performs almost as well as continuous communication.

In summary, this work develops a new framework for discrete communication in multi-agent systems, setting the ground for a broader investigation of emergent artificial communication and facilitating future work on interfacing with these systems.

\section{Background}
\label{sec:bg}
% This work suggests a new discrete communication type, namely quantized communication, and shows that it is significantly superior to known approaches for learning discrete language in an emergent communication setup.
We begin with a formal definition of the multi-agent communication setup, often called ``emergent multi-agent communication'' \cite{lazaridou2020emergent_1}.  %environment and the referential game.
% Next we describe two communication types: Continuous (CN), and Gumbel-softmax (GS), which serve as our baselines.
% Then, in the following section, we describe our suggested quantized (QT) communication.
% 
% \subsection{Emergent Communication via  Games}
% \label{sec:3.emm-comm}
% Under the emergent communication setup, 
In this setup, a sender and a receiver communicate in order to accomplish a given task.
In the \textbf{referential game}, the sender needs to transmit information about a target object, which the receiver uses to identify the object from a set of candidates. In the \textbf{classification game}, the sender again transmits information about an object, but the receiver needs to identify the \emph{class} the object belongs to, rather than its identity. 
Notably, the two games require significantly different communication.
While in the referential game the sender needs to accurately describe the target, in the classification game the sender needs to describe the target's class (see Appendix \ref{sec:app-ref-cls} for details).

Formally, we assume a world $\sO$ with $| \sO |$  objects.\footnote{We defer details on the type of objects to Section \ref{sec:setup}. For now, one can think of objects as images, texts, etc.} At each turn, $n$ candidate objects $\sC = \{c\}^n \subseteq \sO$, are drawn uniformly at random from $\sO$. One of them is randomly chosen to be the target $t$, while the rest, $\sD = \sC \setminus t$, serve as distractors.

\begin{figure}
\begin{center}
% \framebox[4.0in]{$\;$}
\resizebox{8cm}{!} {\includegraphics{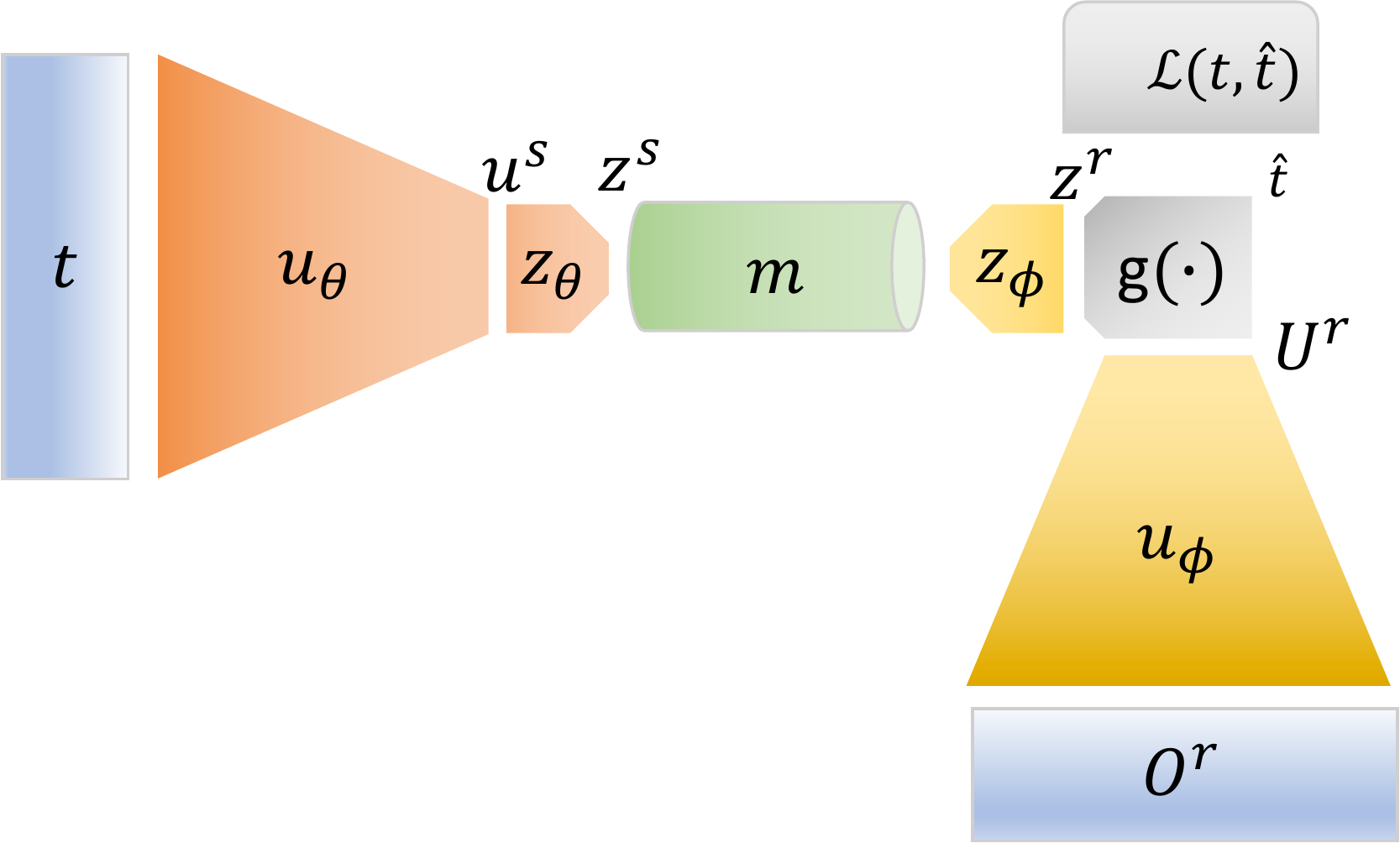}}
% \fbox{\rule[-.5cm]{0cm}{4cm} \rule[-.5cm]{4cm}{0cm}}
\end{center}
\caption{The emergent communication setup.
Sender network is on the left, Receiver network is at the bottom right, and $m$ is the communication channel.}
\label{fig:emm-comm}
\end{figure}

Figure \ref{fig:emm-comm} illustrates the basic setup.
At each turn, the sender $S$ encodes the target object $t$ via its encoder network $u_\theta$, such that  $u^s = u_\theta(t) \in \mathbb{R}^{d}$ is the encoded representation of $t$. 
It then uses its channel network $z_\theta$ to generate a message $m$, 
$m = z_\theta(u^s) = z_\theta(u_\theta(t))$.
The channel and message have certain characteristics that influence both the emergent communication and the agents' performance in the game, and are described in Section \ref{sec:bg-comm-elems}. 

At each turn, the receiver $R$ encodes each candidate object $c_i$ ($i=1,2,\ldots,n$) via its encoder network $u_\phi$, to obtain $u_\phi(c_i)$. 
We write $U^r \in \mathbb{R}^{\sC \times d}$  to refer to the set of encoded candidate representations, each of dimension $d$. 
The receiver then decodes the message via its decoder channel network $z_\phi$, obtaining
$z^r = z_\phi(m) \in \mathbb{R}^d$.
Next, the receiver computes a score matching each of the encoded candidates to the decoded message.
%produces a logits vector $\vg = g(\cdot)$ to decide which of the candidates is the target:
% $g^r(z^r, U^r) \rightarrow \sR^{|\sC|} = \mathrm{softmax}(z^r \cdot U^r)$.
The receiver then calculates prediction scores $\tilde{\vt} =  \mathrm{softmax}(z^r \cdot U^r)$. 
% 
% its target prediction $\hat{t}$ by:
% $\hat{t} = \mathrm{softmax}(z^r \cdot U^r) = \mathrm{softmax}(z_\phi(m) \cdot u_\phi(C)) = f_\phi(m, C)$.
% 
At test time, the receiver's predicted target object is the one with the highest score, namely  $\hat{t} = \mathrm{argmax}_i \tilde{\vt}$. 
During training, the entire system is optimized end-to-end with the cross-entropy loss between the correct target $t$ and the predicted target $\hat{t}$. 
The trainable parameters are all the parameters of both sender and receiver networks, $\theta$ and $\phi$. 

% Finally, the receiver uses the cross-entropy loss $\gL$ to calculate the loss and update network parameters ($\theta_u, \theta_z, \phi_u, \phi_z$) of both sender and receiver networks:
% $\gL = CE(Y, \hat{Y}) = CE(\1_\mathrm{t}, g^r)$, where $\1_\mathrm{t}$ is a one-hot vector of the target's index. 

\subsection{Communication Elements}
\label{sec:bg-comm-elems}
A key aspect of the emergent communication setup is the message ($m$ in Figure \ref{fig:emm-comm}). 
In this work we compare three communication modes that generate this message: continuous (CN) uses a continuous message, while Gumbel-softmax (GS) and quantized (QT) use a discrete message. 
We start by describing the communication elements common to all modes, and then provide more details on the unique aspects of each communication mode.

Formally, we define three communication elements, namely, symbol, word and message. Figure \ref{fig:channel} provides an example for each element.
\begin{itemize}
\item \textbf{Symbol} is the atomic element of the communication.
An alphabet is a collection of symbols.
The alphabet size is a parameter of the quantized and Gumbel-softmax communication modes, while continuous communication uses real-numbers in $m$, corresponding to an uncountable alphabet.

\item \textbf{Word} is the basic message element. A word is represented with a one-dimensional vector.
In continuous communication, this vector is composed of floating point numbers, for quantized communication it is composed of integers, and for Gumbel-softmax communication it is a one-hot vector.

\item \textbf{Message} is a sequence of one or more words, which the sender sends to the receiver. 
An instantaneous (Instant) channel is capable of sending (and receiving) only single-word messages, while a Recurrent channel sends (receives) multi-word messages with a recurrent neural network (RNN). %We use the implementation in Egg \cite{kharitonov2019egg_a20}.
\end{itemize}
 
\subsection{Communication Modes}
In this section we describe two known communication modes: continuous and Gumbel-softmax.
These communication modes serve as baselines.
In the following section we describe our quantized communication.

\subsubsection{Continuous Communication} 
In continuous communication, words are represented with floating point vectors (see Figure \ref{fig:channel}).
Though continuous, one may think of each vector element as if it represents a symbol, and the vector itself represents a word.
Continuous communication is expected to lead to good performance, provided that the channel has sufficient capacity.
With continuous communication, the system can easily be trained end-to-end with back-propagation.
%Such a representation has two clear advantages. First, it provides high channel capacity, and, second, it allows optimization via gradient decent, thus facilitating training the system end-to-end.
%Clearly, continuous communication does not allow a discrete language to emerge, and thus, is not a valid communication option for researching the characteristics of the emergent language.
%Therefore, one must find ways to digitize the continuous communication.
%Still, we use this communication mode as an upper-performance baseline.
%Interestingly, we found during the experiments that our quantized communication sometimes reach even better performance.

\subsubsection{Gumbel-softmax}
The Gumbel-softmax is a continuous approximation for a categorical distribution. In the communication context, a discrete message is approximated via a sampling procedure. Details are given elsewhere \cite{havrylov2017emergence_20,jang2016categorical_a2} and implementation specifics are provided in Appendix \ref{sec:app-hparam}. 
The end result is a continuous message, where each word has the size of the alphabet and holds \emph{one} (approximate) symbol. This allows for end-to-end optimization with gradient methods, and for discrete communication at inference time. However, the channel capacity is limited, and a large alphabet size is both inefficient (due to the need to sample from a large number of categories) and does not perform well in practice.

\section{Quantized Communication}
% \subsection{Quantization}
Quantization techniques aim to reduce model size and computation cost while maintaining a similar level of performance to the original model \cite{banner2018scalable_a9}. 
The key quantization idea is to replace floating-point representations of model weights and/or activations with integers.
We emphasize that, while quantization has a specific purpose in mind (efficiency), it renders the neural network discrete by definition.
Allowing gradients to flow though the network during back-propagation  enables end-to-end gradient-based optimization of the network with off-the-shelf optimizers.

\subsection{Quantized Communication Method}
\label{sec:qt_comm}
We follow the quantization definition and notation provided by \citet{gholami2021survey_102}.
The quantization operator is defined by
$ Q(r) = \mathrm{Int}(r/S) - Z$
where $r$ is a real-valued floating-point tensor, $S$ is a real-valued scaling scalar, and $Z$ is an integer zero point, which we set to zero.
The Int() function maps a real value to an integer value through a rounding operation (e.g., round to nearest integer and truncation).
This operator, also known as \textit{uniform quantization} \cite{gholami2021survey_102}, results in quantized values that are uniformly spaced.\footnote{
Future work may explore communication with non-uniform quantization schemes \cite{gholami2021survey_102}.}
%We leave it for future work to explore and refine the uniform quantization with better techniques. 

One can recover floating-point values $r$ from the quantized values $Q(r)$ through  dequantization,
$\hat{r} = S(Q(r) + Z)$.
Obviously, the recovered real values $\hat{r}$ will not exactly match $r$ due to the rounding operation.
This rounding mismatch is a core difference between continuous and quantized communication.

\begin{algorithm}[t]
\caption{Quantizing continuous communication.}
\label{alg:quantize}
\begin{algorithmic}[1]
\State msg: continuous message \Comment{Floating point vector}
\State S: scaling factor \Comment{Set the alphabet range}
\Procedure{Normalize}{$msg$} %\Comment{Normalize to $0 \leq msg.elems \leq 1$}
    \State $min\_elem \gets min(msg.elements)$
    \State $max\_elem \gets max(msg.elements)$
    \State $msg \gets (msg.elems - min\_elem) / max\_elem$
    \State \textbf{return} $msg$
\EndProcedure
\Procedure{Quantize}{$msg$}
    \State $ msg \gets Normalize(msg)$
    \State $s\_msg \gets msg / S$ \Comment{Scale to range}
    \State $qt\_msg \gets quantize(s\_msg)$ \Comment{Integer vector}
    \State $deqt\_msg \gets dequant(qt\_msg)$ \Comment{Rounded float}
    \State $discrete\_msg \gets deqt\_msg$ \Comment{For logging}
    \State $deqt\_msg \gets deqt\_msg * S$ \Comment{Scale back}
    \State \textbf{return} $(deqt\_msg, discrete\_msg)$
\EndProcedure
\State \textbf{msg} $\gets Quantize(msg)$
\end{algorithmic}
\end{algorithm}

% \textbf{Controlling the alphabet size:}
The quantized operator's scaling factor $S$ essentially divides a given range of real values $r$ into a number of partitions.
Specifically, we define the scaling factor $S$ to be
$S = \frac{[\beta - \alpha]}{|v|}$,
and we set $v$ to be the alphabet size.
In this work we normalize message values to the range $[0, 1]$, thus $\beta = 1, \alpha=0$. 
Epmirically, message normalization improves results for both quantized and continuous communication.
Notably, the rounding error of $\hat{r}$ is linearly correlated with the alphabet size.

This procedure results in a quantization algorithm, presented in Algorithm~\ref{alg:quantize}, which maps each message to a set of symbols from the alphabet.
% Our algorithm for quantizing a continuous message is shown in Algorithm~\ref{alg:quantize}. 
The quantization algorithm allows \textbf{fine-grained control over channel capacity}.
Capacity can be controlled by both the alphabet size and the word length.
The total number of unique words allowed by the channel is given by $\mathrm{channel\textrm{-}capacity} = \mathrm{alphabet\textrm{-}size}^{\mathrm{word\textrm{-}length}}$.

\subsection{Training with Quantization}
Notably, one may choose to apply the quantization algorithm during both training and inference, or only during inference.
Quantization only during inference is similar to the basic Gumbel-softmax setup described above, where training is done with a continuous approximation and inference is discrete. Quantization during training makes the system non-differentiable due to the rounding operation. In this case, we use the straight-through estimator \cite[STE;][]{bengio2013estimating_a13}, which approximates the non-differentiable rounding operation with an identity function during the backward pass. 
This is similar to what is known in the Gumbel-softmax literature \cite{jang2016categorical_a2} as the straight-through option, where the softmax is replaced with argmax during the forward path.

% Similar to GS, applying quantization during training yields mix results, which we report in the "Results" section.

% The presented quantized algorithm uses the most straight-forward quantization approach.
% We plan to research advances and modifications to this algorithm in future work.

\section{Experimental Setup}
\label{sec:setup}

\subsection{Games and Datasets}
\label{sec:datasets}
We run our experiments on four games from three datasets.

\paragraph{Synthetic objects.} This dataset is based on Egg's object game \cite{kharitonov2019egg_a20}.
Each object has 4 attributes.
Each attribute has 10 possible values, and different attributes share the same set of values.
Thus, the dataset contains $10^4$ objects which are uniquely identified by four discrete values.

\paragraph{Images.}  We use the Egg implementation of the image game from \citet{lazaridou2016multi_14}.\footnote{\url{https://dl.fbaipublicfiles.com/signaling_game_data}}  
The Dataset contains images from ImageNet \cite{deng2009imagenet_a21}.
The training set contains $46,184$ images, distributed evenly over $463$ classes, out of which we randomly choose $8032$.
The validation and test sets have $67,521$ images each, split over the same classes.
We randomly choose distractors from classes other than the targets'.

\paragraph{Texts.}  We use a short text dataset, named Banking77 \cite{casanueva2020efficient_a15}, which we refer to as Sentences.
It contains $10,000$ sentences, classified into 77 classes, each represented with a meaningful class name.
The sentences are user queries to an online customer support banking system, while the classes are user intents.
% Text-based dataset are not commonly used for emergent communication experiments, as it is already based on human language representation.
%Still encoding text (e.g., words and sentences) into dense representations, make it suitable for demonstrating communication aspects in an emergent communication setup.
We use the Sentences dataset for two different games:
\textbf{Sent-Ref} is a referential game, that is, the receiver needs to identify the sentence. % in which we ignore the sentence's label.
%Sentence-Ref is, thus, similar to Object and Image games in the sense that the sender sends a message that describes an encoded sentence which the receiver then needs to identify from a set of candidates.
\textbf{Sent-Cls} is a classification game, where the 
 receiver receives a set of candidate classes and needs to identify the class of the target sent by the sender.  

\bigskip 

\subsubsection{Data Splits.}  In all experiments we split the data 80/10/10 into training, validation, and test sets,  respectively. For Image and Sentence-Ref games, both targets (sender-side objects) and candidates (receiver's side) are mutually exclusive across splits. For Object and Sentence-Cls game, targets are mutually exclusive while candidates are shared across splits.
Table~\ref{tbl:datasets} provides summary statistics of the datatsets. 

\begin{table}[h!]
\centering
\resizebox{\columnwidth}{!}{
%\small
\begin{tabular}{l rrrrr}
\toprule
Dataset & \#Objects & \#Train & \#Valid & \#Test & Max  \\ 
\cmidrule(lr){1-1} \cmidrule(lr){2-6}
Object & 10K & 8000 & 1000 & 1000 & 10K  \\ 
Image & 181K & 8032 & 1024 & 1024 & 100 \\ 
Sent-Ref & 10K & 7997 & 1001 & 1001 & 77 \\ 
Sent-Cls & 10K/77 & 7953 & 1004 & 1042 & 77 \\ \bottomrule
\end{tabular}
}
% \vspace{-0.1in}
\caption{Sizes of datasets and splits for each game.
77 is the number of classes in the Banking dataset.
% For the Image dataset we used a subset of the data.
Max is the maximum number of candidates used for evaluating the game.}
%\end{center}
\label{tbl:datasets}
\end{table}

\subsection{Agents' Architecture}

\subsubsection{Encoding Agents.}
We refer by sender and receiver encoding agents to the $u_{\theta}$ and $u_{\phi}$ networks, respectively, as described in Section \ref{sec:bg}.
For Object and Image games, we follow the architecture provided by the Egg implementation \cite{kharitonov2019egg_a20}.
The agents in the Object game uses a single fully-connected (FC) layer to encode the objects.
The agents in the Image game use a FC network followed by two convolutional layers and a second FC layer.
In the Image game, the sender uses all candidates for encoding the target (referred to as `informed-sender' by  \citet{lazaridou2016multi_14}). In all other games the sender encodes only the target object. 
For the Sentence games (both referential and classification), we use a distilbert-base-uncased model \cite{sanh2019distilbert_22} from Huggingface \cite{wolf-etal-2020-transformers} as the sentence encoder without any modification.\footnote{% 
Importantly, in this work we aim to evaluate communication performance across various settings, and not necessarily find the best-performing encoding network. Nevertheless, we find our setup to achieve close to state-of-the-art results in the Sentence-Cls game, as shown in Section \ref{sec:results}.} 
Appendix \ref{sec:app-hparam} provides more details. 

\subsubsection{Communication Channels.}
We refer by sender and receiver channels to the $z_{\theta}$ and $z_{\phi}$ networks, respectively, as described in Section \ref{sec:bg}.
We experiment with two architectures for the communication channel: Instant and Recurrent.
\textbf{Instant} simply passes the sender's encoded representation of the target  through a FC network to scale it to the word length and send it to the receiver.
%For GS communication mode, the Gumbel softmax is added at this point. {\color{blue} consider adding a formal description here}.
%For quantized communication, quantization is done on the logits by using the quantized algorithm described and section \ref{alg:quantize}.
The receiver's Instant channel decodes the message with a FC feed-forward network and compares it with the candidates' encoded representations as described in Section \ref{sec:bg}.
The \textbf{Recurrent} channel enables sending and receiving multi-word messages.
We adapt the Recurrent channel implemented in Egg to work with continuous and quantized communication.
%Similar to the Instant channel, GS communication add the Gumble layer for every word during the recurrent process. 
%The quantize communication quantize the whole message after sender's channel completed to generate it.
%The Receiver's channel uses a recurrent cell to decode the message.
%It is based on the Egg implementation and is identical for all communication types.
More details on channel configuration are provided in Appendix \ref{sec:app-hparam}

\subsection{Number of Candidates and Distractors}
Most earlier emergent communication setups use a limited number of distractors \cite{mu2021emergent_55, li2019ease_53}.
Recent work \cite{chaabouni2021emergent_100,  guo2021expressivity_81} reports the effect that an increased number of distractors has on accuracy results.
The number of distractors affects results in two complementary ways.
On the one hand, adding more distractors renders the receiver's task harder during inference.
On the other hand, during training, distractors serve as negative samples which are known to improve learning \citet{mitrovic2020less_85}.
%\boaz{another important statement from 100: Note that, in the literature, the number of distractors rarely exceeds a dozen both at train and eval times (Mu \& Goodman, 2021; Li \& Bowling, 2019).}
%Our work, as well as others {\color{blue} ref needed}, show that increasing the number of the negative samples improves results during inference.  
Based on these observations, our experimental environment lets us decouple the number of negative examples during training, from the number of distractors used for evaluation.
In all our experiments we train the system with a large number of negative samples (serving as distractors) and report results on an increasing number of candidates, always including the target as one of them.
%We apply n-batch negatives similar to \citet{guo2021expressivity_81} for the Sent-Ref and Sent-Cls games in order to reduce memory usage.

%\subsubsection{In Batch Contrastive Loss}
%We apply in-batch contrastive learning {\color{blue} ref to in batch contrastive} for the Sentence-ref and Sentence-Cls games due to the large representation objects in this game have. 

\subsection{Evaluation Metrics}
In this work we report prediction accuracy as the main metric.
Similar to \citet{guo2021expressivity_81}, we observe a correlation between the number of unique messages (NoUM) and accuracy, and report this measurement as well. Recent work \cite{chaabouni2021emergent_100, yao2022linking_94} reports that the popular topographic similarity metric \cite{brighton2006understanding_a17, lazaridou2018emergence_41} does not correlate well with accuracy, especially when measured on a large number of distractors.
In our work we observed the same effect, so we refrain from reporting this metric.

\subsection{Training Details}
% We run all experiments with training, validation and test sets. 
% We use early-stopping with patience.
% We report results using a learning rate of 1e-3 for the image game and 1e-5 for all other games.
%Each experiment took less than 24 hours on a single v100 GPU and a CPU with 128GB of RAM. 

We performed hyper-parameter tuning on the validation set and report results on the test set.
As systems have many hyper-parameters to tune we tried to reduce changing most parameters between different setups to a minimum.
However, we ran an extensive hyper-parameter search for alphabet size and message length for the Gumbel-softmax communication to insure that we report the best possible results for this communication mode.
Quantized communication required only minimal tuning and still outperformed Gumbel-softmax across all setups.
We report more details on configurations and hyper-parameters in Appendix \ref{sec:app-hparam}.

Each experiment took under 24 hours on a single v100 GPU and a CPU with 128GB RAM. 
We run each experiment three times with different random seeds and report average performance. Variance is generally small (Appendix \ref{sec:app-results}). 

%Specifically, we experimented and reported extensive search of alphabet size and message length for quantized communication over the Object game when using Recurrent channel with maximum message length of 6. See table \ref{tbl:vocab-word} for details.

%\subsubsection{Quantize during training}
%Quantization process can take place just at inference time or both at training and inference.
%Each of these approaches has its own advantages.
%We experienced with both approaches and realise that their performance are game dependent. More work is needed to better understand the factors that affect model performance for each of these approaches.

\section{Results} \label{sec:results}
\begin{figure}[t]
\centering
% \framebox[4.0in]{$\;$}
\resizebox{8cm}{!} {\includegraphics{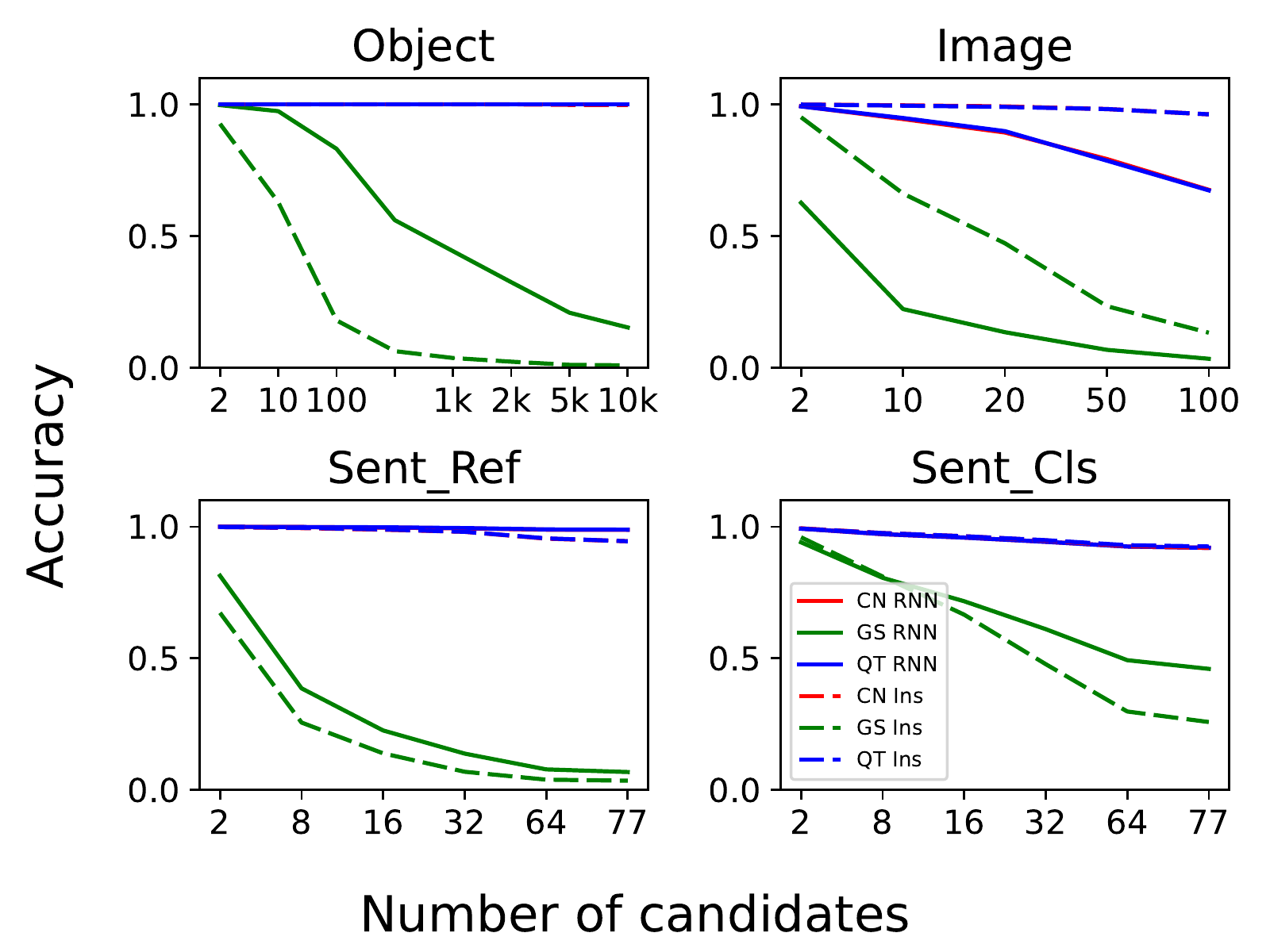}}
% \fbox{\rule[-.5cm]{0cm}{4cm} \rule[-.5cm]{4cm}{0cm}}
\caption{Communication results for four games for Instant (Ins) and Recurrent (RNN) channels, using quantization only during inference.
% The channel is configured with word length of 100 for CN and QT apart from the QT-Rnn of Sent-Cls which has a word length of 10. Alphabet size for QT is set to 10 for all configurations. Alphabet size for GS is set to 10, 50, 100, and 10 for the Rnn channel, and 100, 50, 100, and 100 for the Instant channel for the Object, Image, Sent-Ref and Sent-Cls games, respectively. %which in this case equals W-len, and both W-len and V-size for the QT communication.
%Maximum message length is six for all Rnn configurations.
%We aim to maintain the same channel parameters across communication types.
The CN and QT results are essentially the same, thus overlapping in the plots.
Channel parameters for the experiments are provided in footnote \ref{fn:main-config}.
% %\ref{tbl:main-obj}-\ref{tbl:main-set-cls}
}
\label{fig:main-graph}
\end{figure}

We first experiment with quantization only during inference and compare its performance to continuous and Gumbel-softmax communication. 
Our main results are presented in Figure \ref{fig:main-graph}.\footnote{These results are with the best-tuned configurations: 
Word  length of 100 for continuous (CN) and quantized (QT) modes, except for QT-RNN in Sent-Cls, where word length is 10. Alphabet size of QT is 10 in all configurations. For Gumbel-softmax (GS),  Alphabet size is 10, 50, 100, and 10 for the RNN channel, and 100, 50, 100, and 100 for the Instant channel, for the Object, Image, Sent-Ref, and Sent-Cls games, respectively.
Section \ref{sec:results-lang} and Appendix \ref{sec:app-lang} provide results with a range of possible configurations.}
\label{fn:main-config}
The graphs show test accuracy results (Y-axis) against the number of candidates (X-axis) for the three communication modes over two channel architectures for the four games.
As expected, performance generally degrades when increasing the number of candidates in most setups. This is especially evident in Gumbel-softamx (GS) communication (green lines), while continuous (CN, red) and quantized (QT, blue) communication modes scale much more gracefully with the number of candidates. Notably, the \textbf{quantized communication is on-par with the fully continuous communication in all cases}.

Considering the different games, in the Object game, continuous and quantized communication perform perfectly, while Gumbel-softamx suffers from a large number of candidates. In the other games, there is a slight performance degradation  with continuous and quantized communication. 

Next we compare the performance with communication using Instant vs.\ Recurrent channels. Recall that the Instant channel has a much more limited capacity (each message is one-word long) compared to the Recurrent channel (each message is made of multiple words). 
The Gumbel-softmax communication suffers most clearly from the Instant channel: In all but the Image game, it performs worse than Gumbel-softmax with the Recurrent channel. The gap is especially striking in the Object game (compare green dashed and solid lines).  
The poor performance of Gumbel-softmax with the Instant channel can be explained by the limited channel capacity: The sender in this case can generate up to 100 unique messages (each containing just a single symbol), which are just 1.0\% of the unique objects in the Object game. Thus it has little hope of correctly identifying the target object. 
One might hope that a Recurrent channel would help Gumbel-Softamx perform better, as it has the required capacity to uniquely represent all objects ($10^6$ unique messages). 
However, even with this capacity, performance on a high number of candidates is low. We attribute the poor performance to the difficulty to optimize RNNs with discrete communication. This might also explain why Gumbel-softmax with an instant channel works better than the one with Recurrent channel in the Image game. 

In contrast to Gumbel-softmax, quantized communication does not suffer from the limited capacity problem, nor from the optimization difficulty. In both Instant and Recurrent channels, quantized communication leads to excellent performance, even in the face of a large number of candidates.

We note in passing, that for the sentence classification game (Sent-Cls), we get results that are on par with state-of-the-art classification results for this dataset \cite{qi2020benchmarking_a18, zhang2021few_a19}, even though we use a very different setup, that of communication between agents.

\subsection{Quantization During Training}
So far we report results when quantization is applied only at inference time. 
Here we compare it with applying quantization also during training. 
Table \ref{tbl:qt-training} reports accuracy results for the two settings, using either two or a maximum number of candidates (varying by game). 

As seen, performance results are on-par for all games and all communication settings, whether using quantization during training and inference or only at inference.
The quantized communication achieves perfect, or near-perfect accuracy ($>0.99\%$) for setups with 2 candidates.
Accuracy results surpass 92\% for all games, apart from the Image game, even when the receiver has to discriminate between the maximum number of candidates.
The low Image game results are attributed to the use of suboptimal convolutional networks at both the sender and receiver.

%In the Object game, we report perfect performance in all communication settings, whether using quantization in training and inference or only at inference. 
% In the other games, quantization only during inference leads to better performance than quantization during both training and inference. This may be explained by the added constraint of learning to perform a task with genuine discrete communication. However, it is striking that even with quantization during training, results are quite good in many cases, especially with the Recurrent channel. We conclude that quantization is a beneficial approach for learning to communicate with a discrete channel. Note that since we used a very naive quantization scheme during training, it is likely that more elaborate discretization schemes, e.g., non-uniform \cite{choi2018bridging_a16}, may greatly improve performance.   

\begin{table}[t]
\centering
\resizebox{\columnwidth}{!}{
%\small
\begin{tabular}{l rr rr rr rr}
\toprule 
& \multicolumn{4}{c}{Instant} & \multicolumn{4}{c}{Recurrent} \\ 
\cmidrule(lr){2-5} \cmidrule(lr){6-9} 
& \multicolumn{2}{c}{Train+Inf} & \multicolumn{2}{c}{Only Inf} & \multicolumn{2}{c}{Train+Inf} & \multicolumn{2}{c}{Only Inf} \\ 
\cmidrule(lr){2-3} \cmidrule(lr){4-5} \cmidrule(lr){6-7} \cmidrule(lr){8-9} 
%  & \multicolumn{2}{c|}{Inst QT-Train} & \multicolumn{2}{c|}{Inst CN-Train} & \multicolumn{2}{c|}{Rnn QT-Train} & \multicolumn{2}{c}{Rnn CN-Train} \\ 
Game & \multicolumn{1}{c}{2} & \multicolumn{1}{c}{Max} & \multicolumn{1}{c}{2} & \multicolumn{1}{c}{Max} & \multicolumn{1}{c}{2} & \multicolumn{1}{c}{Max} & \multicolumn{1}{c}{2} & \multicolumn{1}{c}{Max} \\ 
\cmidrule(lr){1-1} \cmidrule(lr){2-9}
Object   & 1.00 & 1.00\hspace{0.20cm} & 1.00 & 1.00 & 1.00 & 1.00\hspace{0.17cm}     & 1.00 & 1.00 \\ 
Image    & 1.00 & 0.96\hspace{0.20cm} & 0.99 & 0.96 & 0.99 & 0.68\textsuperscript{*} & 0.99 & 0.67 \\ 
Sent-Ref & 1.00 & 0.95\hspace{0.20cm} & 0.99 & 0.95 & 1.00 & 0.99\hspace{0.17cm} & 1.00 & 0.99 \\ 
Sent-Cls & 0.99 & 0.92\hspace{0.20cm} & 0.99 & 0.92 & 0.99 & 0.92\hspace{0.17cm}     & 0.99 & 0.93 \\ \bottomrule
\end{tabular}
}
%\vspace{-0.1in} - vspace not allowed
\caption{Reporting quantization during both training and inference (Train+Inf) and quantization only during inference (Only Inf), for Instant and Recurrent channels, with 2 candidates and the maximum number of candidates in each game (Max).
Reported results are for an average of three runs. Standard deviations for all reported results are under $0.01$, apart from the one marked by $*$, which is under $0.05$.
}
%\end{center}
\label{tbl:qt-training}
\end{table}

\subsection{Communication Analysis}
\label{sec:results-lang}
\begin{figure}[t]
\centering
\includegraphics[width=1.0\columnwidth]{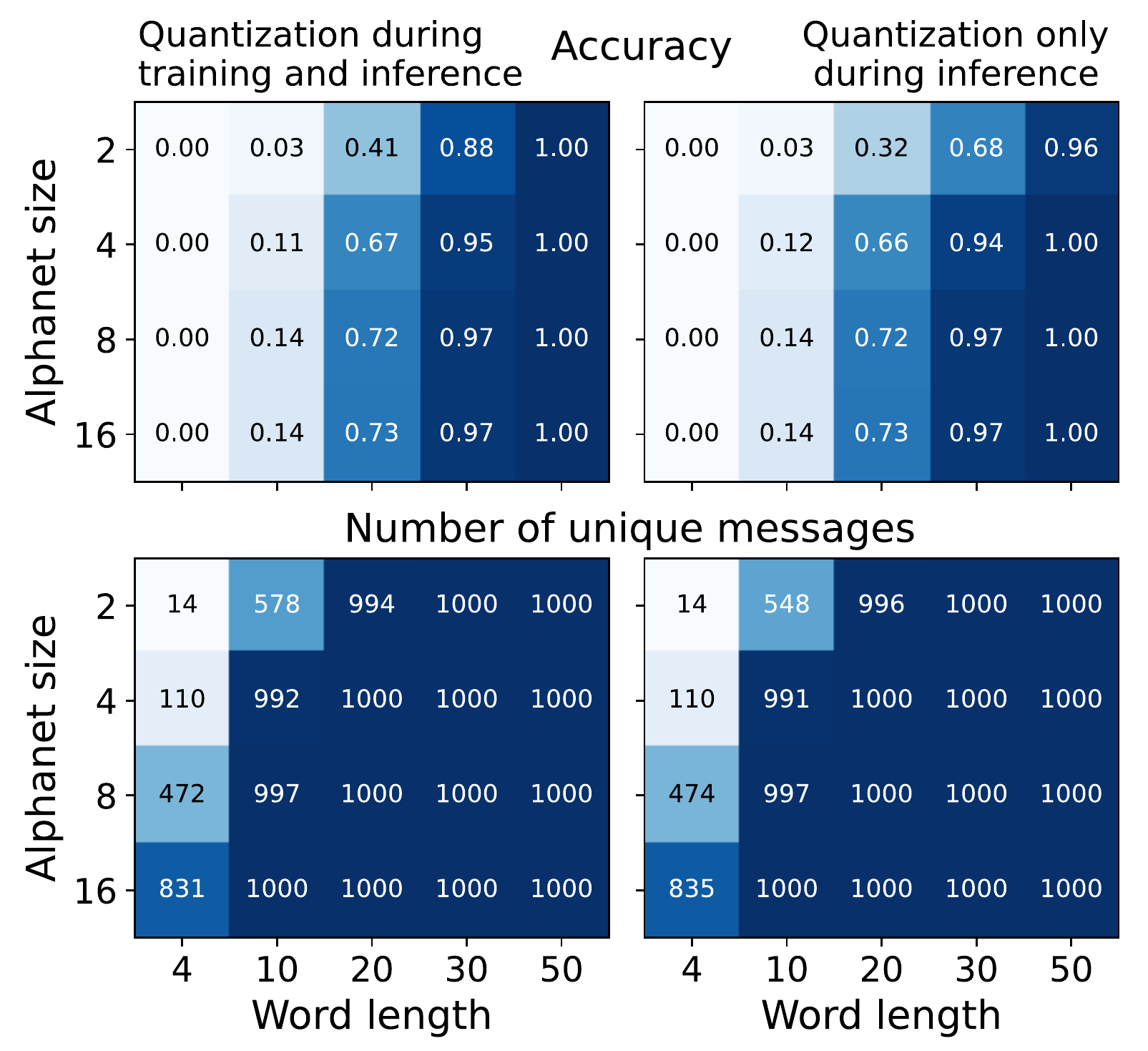}
\caption{Showing accuracy (top) and number of unique messages (NoUM) (bottom) as a function of alphabet size and word length for the Object game with an Instant channel.
Comparing quantization during both training and inference (left) or only during inference (right).}
\label{fig:lang_analysis}
\end{figure}

Figure \ref{fig:lang_analysis} analyzes the quantized communication results for the Object game over an Instant channel.
Results are obtained from a test set with 1000 unique targets.
Appendix~\ref{sec:app-lang} provides a similar analysis for the other games, showing largely consistent results. 

The top heatmaps show performance in various settings, organized according to word length (X-axis) and alphabet size (Y-axis).
The bottom heatmaps show the number of unique messages (NoUM) sent by the sender in each configuration (1000 max).
We compare quantization during both training and inference (left heatmaps) with quantization only during inference (right heatmaps).
As seen, quantization during both training and inference performs slightly better than quantization only during inference. 

% The last row indicates results for continuous communication.
Clearly, increasing the channel capacity (moving to the bottom-right corner in the heatmaps) improves accuracy results and increases the NoUM sent by the sender up to a maximum of 1000.
Increasing word length (moving to the right) improves results substantially for all alphabet sizes, and reaches a maximal performance for a length of 50, for all alphabet sizes, when quantization is done during both training and inference, and for alphabet size larger than 4 for quantization during inference only (top right).
%For quantization during both training and inference, the same results are achieved with an alphabet size larger than 16 (top left).
Interestingly, increasing the alphabet size (moving to the bottom in the heatmaps) has a smaller effect.
With long enough words, the system performs almost optimally, even with a very small alphabet (e.g., 2 and 4 symbols), resembling findings by \citet{freed2020communication_105}. 
As seen by comparing the top and bottom heatmaps, the NoUM correlates with performance.  
Interestingly, having a NoUM equal to the number of unique targets is a necessary but not sufficient condition for perfect performance.

% Results do not significantly improved for most word lengths, except for word length of 20, for which results for the 10K candidates improves form 10\% to 90\% for alphabet of size one (2 symbols) and 255 (2256 symbols), respectively.
% Looking further, one may observe that average message length is higher for channel with limited capacity, probably due to the need to compensate for this.
% As seen by the 5 upper-left cells in the table, this compensation is not optimal as the channel is still not able to generate enough unique messages for describing all 1000 objects. 
%\begin{figure}[h]
%\begin{center}
% \framebox[4.0in]{$\;$}
%\resizebox{8cm}{!} {\includegraphics{quantized-communication/figures/lang_analysis_with_len.pdf}}
% \fbox{\rule[-.5cm]{0cm}{4cm} \rule[-.5cm]{4cm}{0cm}}
%\end{center}

Finally, we compare the number of unique messages in Gumbel-softmax and quantized communication modes across the four games (Table~\ref{tbl:noum_gs_qt}).
The number of unique messages generated by quantized communication equals (in the Object and Sent-Ref games) or almost equals (in the Image and Sent-CLS games) the number of unique targets. In contrast, Gumbel-softmax does not generate enough unique messages.
The Gumbel-softmax with a Recurrent channel produces many more messages than the Instant channel. However, only for the Object game does it generate nearly enough unique messages.
It is noteworthy that sender at the Sent-Cls game does not require to generate unique message for every target for optimally solving it.

\begin{table}[t]
\centering
\begin{tabular}{l c r r r r}\toprule 
&  & \multicolumn{2}{c}{Instant} & \multicolumn{2}{c}{Recurrent} \\ 
\cmidrule(lr){3-4} \cmidrule(lr){5-6} 
Game & \#Targets & GS & QT & GS & QT\\ \cmidrule(lr){1-1} \cmidrule(lr){2-6}
Object & 1000 & 58 & 1000 & 948 & 1000\\
Image & 1024 & 16 & 1016 & 391 & 1016 \\
Sent-Ref & 1001 & 8 & 1001 & 116 & 1001\\
Sent-Cls & 1042 & 20 & 1042 & 125 & 1035\\\bottomrule
\end{tabular}
\caption{Number of test-set targets and unique messages for GS and QT communication modes, with Instant and Recurrent channels.
%Alphabet size for GS is set to 10, 50, 100, and 10 for the Recurrent channel, and 100, 50, 100, and 100 for the Instant channel for the Object, Image, Sent-Ref and Sent-Cls games, respectively.
Channel parameters are provided in fn.\ \ref{fn:main-config}.
% QT alphabet size is 10 and word length is 100 for all games.
% QT results are for quantization only during inference.
}
\label{tbl:noum_gs_qt}
\end{table}

\section{Related work}
The field of emergent communication gained renewed interest in recent years with the advances of deep neural networks and natural language processing \cite{lazaridou2020emergent_1}.
Despite significant advances, approaches for generating discrete messages from neural networks remain scarce.
Multi-agent reinforcement learning (RL) and Gumbel-softmax (GS) are the two alternative approaches used by the community.
% Each of these has its own advantages and disadvantages.
% Both provide a limited (i.e., binary) channel capacity while discretizing a continuous message.

\subsection{Multi-Agent Reinforcement Learning}
The work by \citet{foerster2016learning_10} is probably the first to suggest methods for learning multi-agent communication.
%1
%This work proposed two related approaches, know as Reinforced Inter-Agent Learning (RIAL) and Differentiable Inter-Agent Learning (DIAL).
Many studies in the emergent communication field \cite{lazaridou2020multi_29} use RL and variants of the  REINFORCE algorithm \cite{williams1992simple_a4} for solving the referential game \cite{foerster2016learning_10, lazaridou2016multi_14, chaabouni2021emergent_100}.
\citet{vanneste2022analysis_103} provide a comprehensive review of the various ways to overcome the discretization issue within multi-agent environments. 
%Several approaches where suggested along recent years to overcome the discretization issue in a reinforcement learning environment \citep{vanneste2022analysis_103, kharitonov2020entropy_63}.  
Notably, they find that none of the surveyed methods is best in all environments, and that the optimal discretization method greatly depends on the environment.
Somewhat close to our approach, \citet{freed2020communication_105} 
propose an elaborate stochastic quantization procedure, which relies on adding stochastic noise as part of an encoding/decoding procedure, and evaluate it in path-finding and search problems.  
In contrast, our approach is simple and deterministic, and works exceptionally well in the referential and classification games. 
%which makes the communication channel mathematically equivalent to an analog channel with additive noise, through which gradients can be backpropagated.
% In contrast, our approach does not rely on adding noise to the channel, and is capable of transmitting multi-valued symbols within each vector element. 
%As another example, recent work by \cite{chaabouni2021emergent_100} uses RL for analysing emergent communication at scale.
%They reports the difficulty to optimize the system when using large number of distractors.
%They suggest KL regularization to stabilize and improve training.
%With quantized communication these optimization issues do not accrue.   

\subsection{Gumbel-softmax Communication}
Gumbel-softmax \cite{jang2016categorical_a2} enables discrete communication by sampling from a categorical Gumbel distribution.
It allows gradients to flow through this non-differentiable distribution by replacing it with a differentiable sample from a Gumbel-softmax distribution. 
 \citet{havrylov2017emergence_20} compare communication with RL and Gumbel-softmax and observe that the latter converges much faster  and results in more effective protocols.
Since then, many studies have used Gumbel-softmax in the emergent communication setup  \cite{resnick2019capacity_75, guo2021expressivity_81, mu2021emergent_55, dessi2021interpretable_62}.
% Boaz - 15/1/2023 the below sentence was added from space perspective
as it easier to work with than RL-based methods and can be trained end-to-end with gradient decent and back-propagation.
Though widely used as the default method for overcoming the discretization difficulty, Gumbel-Softmax still suffers from at least two severe limitations.
First, it uses a one-hot vector to encode symbols, limiting capacity. Second, it requires sampling from a distribution, making optimization more expensive and less accurate.

\subsection{Quantization}
In the context of neural networks, quantization is a method for reducing model size and computation cost while maintaining performance.
More generally, quantization, as a method to map  input values in a large (often continuous) set to output values in a small (often finite) set, has a long history \cite{gray1998quantization_a5}.
%2
%The history of the theory and practice of quantization dates to 1948, although similar ideas had appeared in the literature as long ago as 1898 \cite{gray1998quantization_a5}.
The fundamental role of quantization in modulation and analog-to-digital conversion was first recognized during the early development of pulse-code modulation systems, especially in the work of \citet{oliver1948philosophy_a6}, and with the seminal coding theory work by \citet{shannon1948mathematical_a7} that present the quantization effect and its use in coding theory. 
% At 1948, with the advent of the digital computer, Shannon wrote his seminal paper (\cite{shannon1948mathematical_a7}), that present the effect of quantization and its use in coding theory.
%3
%Quantization appears in a slightly different way in algorithms that use numerical approximation for problems involving continuous mathematical quantities \cite{trefethen1997numerical_a8}. 
Recently, intensive research on quantization shows great and consistent success in both training and inference of neural networks using 8-bit number representations, and even less \cite{banner2018scalable_a9, wang2018training_a10, choi2018bridging_a16}. 
In particular, breakthroughs of half-precision and mixed-precision training \cite{courbariaux2014training_a11, gupta2015deep_a12} significantly contributed to vast performance improvements.
Notably, moving from floating-point to integer computation renders many operations non-differentiable.
To overcome this subtlety, a straight-through estimator (STE) \cite{bengio2013estimating_a13} is often used.
The STE approximates the non-differentiable rounding operation with an identity function during back-propagation, thus enables end-to-end model training.
%In this work we utilize the rounding operation, described in the "Quantized Communication" section with an API provided by Pytorch. This API use STE during back-propagation, thus allow the model to be trained end-to-end.

\section{Conclusions}
Research on emergent communication between artificial agents strives for discrete communication. However, common methods such as continuous relaxations via Gumbel-softmax lag far behind continuous communication in terms of performance on the agents' task. 
In this work we propose an alternative approach that achieves discrete communication via message quantization, while enabling simple end-to-end training. 
We show that our quantized communication allows us to run the gamut from continuous to discrete communication by controlling the quantization level, namely, the size of the used alphabet and the word length.
When applying quantization we observe extremely good results, even for the smallest possible alphabet size, given long enough word length.
%We also investigated the option to apply quantization during training, demonstrating superior performance compared to the common Gumbel-softmax approach, albeit inferior to inference-only quantization.

Future work may explore more elaborate quantization schemes for message discretization, during either training or inference. 
We believe that the quantization approach offers a good test bed for investigating emergent communication in multi-agent systems.
% In this work we use a uniform quantization scheme during training, and a local message normalization.
% It is likely that more elaborate, non-uniform, discretization schemes, and global message normalization techniques improve performance significantly.
% We leave this investigation to future research.

%Achieving good performance on a task is a precondition for meaningful communication.
%Thus, in future research we plan to show that the generated communication better correlates with human language; an aspect that with current communication modes was shown with just minimal success.

\section*{Acknowledgements}
The work of RM was partially supported by the Skillman chair in biomedical sciences, and by the Ollendor Center of the Viterbi Faculty of Electrical and Computer Engineering at the Technion. 
The work of YB was partly supported by the ISRAEL SCIENCE FOUNDATION (grant No.\ 448/20) and by an Azrieli Foundation Early Career Faculty Fellowship.

% Use \bibliography{yourbibfile} instead or the References section will not appear in your paper
%\bibliography{quantized-communication/qt.bib}
\bibliography{main.bib}

% \bibliography{iclr2022_conference}
% \bibliographystyle{iclr2022_conference}

\clearpage 

\appendix
\appendix

\section{Appendix}
\label{sec:appendix}

\subsection{Detailed results}
\label{sec:app-results}
 Tables \ref{tbl:main-obj}, \ref{tbl:main-image}, \ref{tbl:main-sent-ref}, and \ref{tbl:main-set-cls} provide results for the Object, Image, Sentence-referential, and Sentence-classification games, respectively, which complement the main results from Section \ref{sec:results}.
Each table contains results for six configurations: three communication modes (continuous, Gumbel-softmax, and quantized) and two channel types (Instant and Recurrent). 
The tables also specify the channel settings (alphabet size, word length, and message length) in each configuration. 

The tables provide average results and standard deviations over three runs with different random seeds. Variance is generally small, with the exception of Gumbel-softmax with a Recurrent channel, which exhibits larger variance, attesting to the optimization difficulties in this setup.

\subsection{Referential and Classification Games}
\label{sec:app-ref-cls}
While in the referential game the receiver and sender share objects form the same world, in the classification game they do not.
In the classification game, each object  $o = (o_s, o_r)$ is composed of two elements: a sample $o_s \in \sO_s$ and a label $o_r  \in \sO_r$. 
The sender's targets are drawn from the set of possible samples, $\sO_s$, while the receiver's candidates are drawn from the set of possible labels, $\sO_r$.
For example, in our experiments with the Sent-Cls dataset, the samples are user queries and the labels are their intents. 
Importantly, while the sender's objects are unique, %namely $| \sO_s | = | \sO |$, 
the receiver's objects are not;
% , namely $| \sO_r | \ll | \sO |$; 
for example, two user queries may have the same intent.  The game's goal is to match the sender's target $o_s$ with the receiver's target $o_r$.

Two complementary differences exist between Referential and Classification games. 
On the one hand, it is easy to notice that with continuous communication a trivial solution exists for referential games, given that:
(1) The same encoder is used by the sender and the receiver to encode the target and the candidates, respectively.
(2) The channel has enough capacity to transmit the encoded target within a single message.
Assuming these two conditions hold, the sender can simply send the encoded target to the receiver.
The receiver then just needs to compare the target with its candidates to find the correct one.
Thus, one can expect perfect performance in these cases.

The classification game, on the other hand, requires mapping $|\sO_s|$ objects to $|\sO_r|$ classes. 
Thus, the task can be solved by sending just $|\sO_r|$ unique messages, one for each class.
Still, messages are very different from the objects they describe, thus continuous communication does not have a foreseen advantage over other communication types in the Classification game, beyond those related to optimization differences.

\subsection{Language Analysis for the Three Games}
\label{sec:app-lang}
\begin{figure}[h]
\centering
\includegraphics[width=1.0\columnwidth]{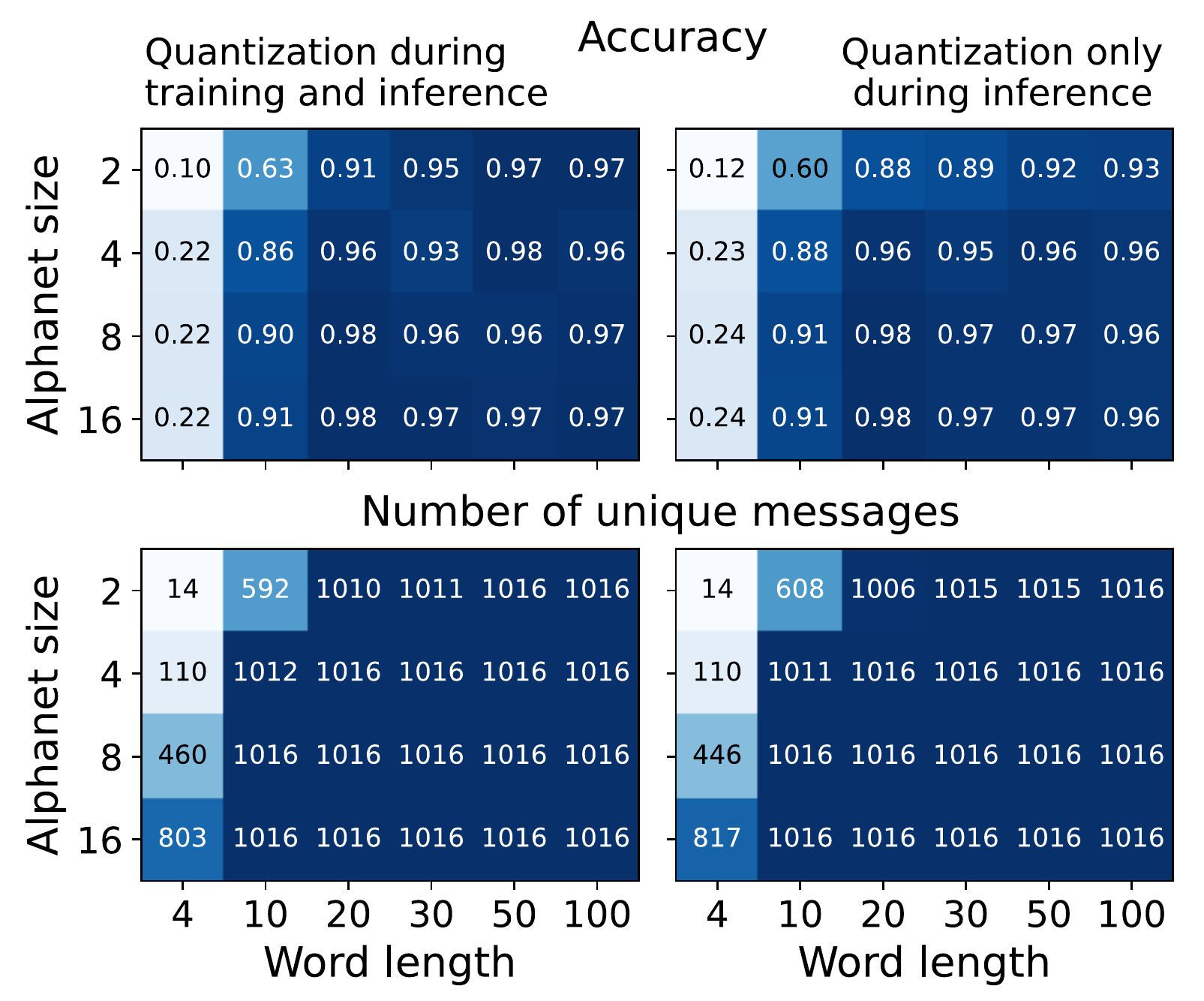}
\caption{Accuracy (top) and number of unique messages (NoUM) (bottom) as a function of alphabet size and word length for the \textbf{Image} game with an Instant channel, comparing  quantization during both training and inference (left) or only during inference (right).}
\label{fig:image_lang_analysis}
\end{figure}

\begin{figure}[h]
\centering
\includegraphics[width=1.0\columnwidth]{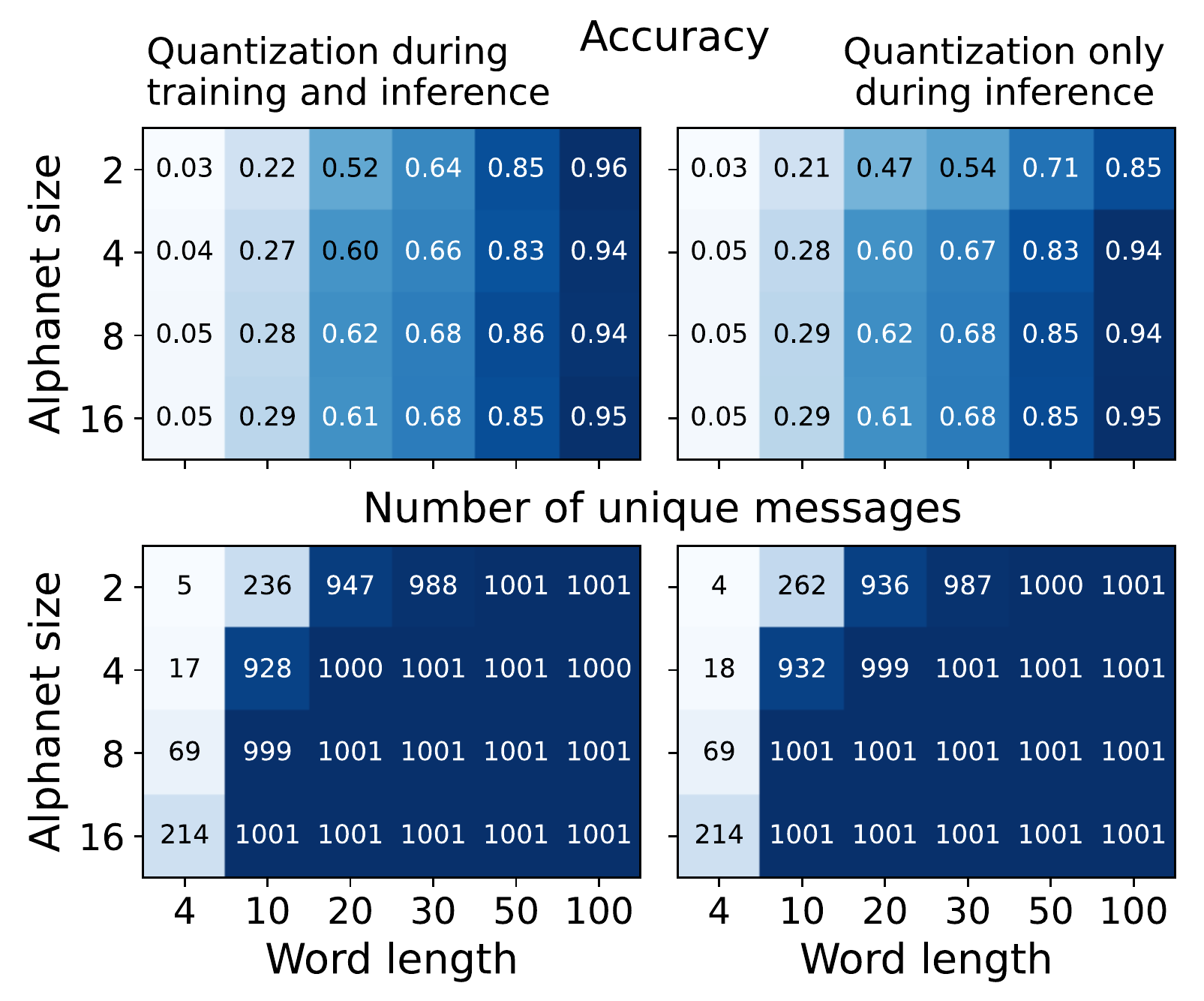}
\caption{Accuracy (top) and number of unique messages (NoUM) (bottom) as a function of alphabet size and word length for the \textbf{Sent-Ref} game with an Instant channel, comparing  quantization during both training and inference (left) or only during inference (right).}
\label{fig:sent_ref_lang_analysis}
\end{figure}

\begin{figure}[h]
\centering
\includegraphics[width=1.0\columnwidth]{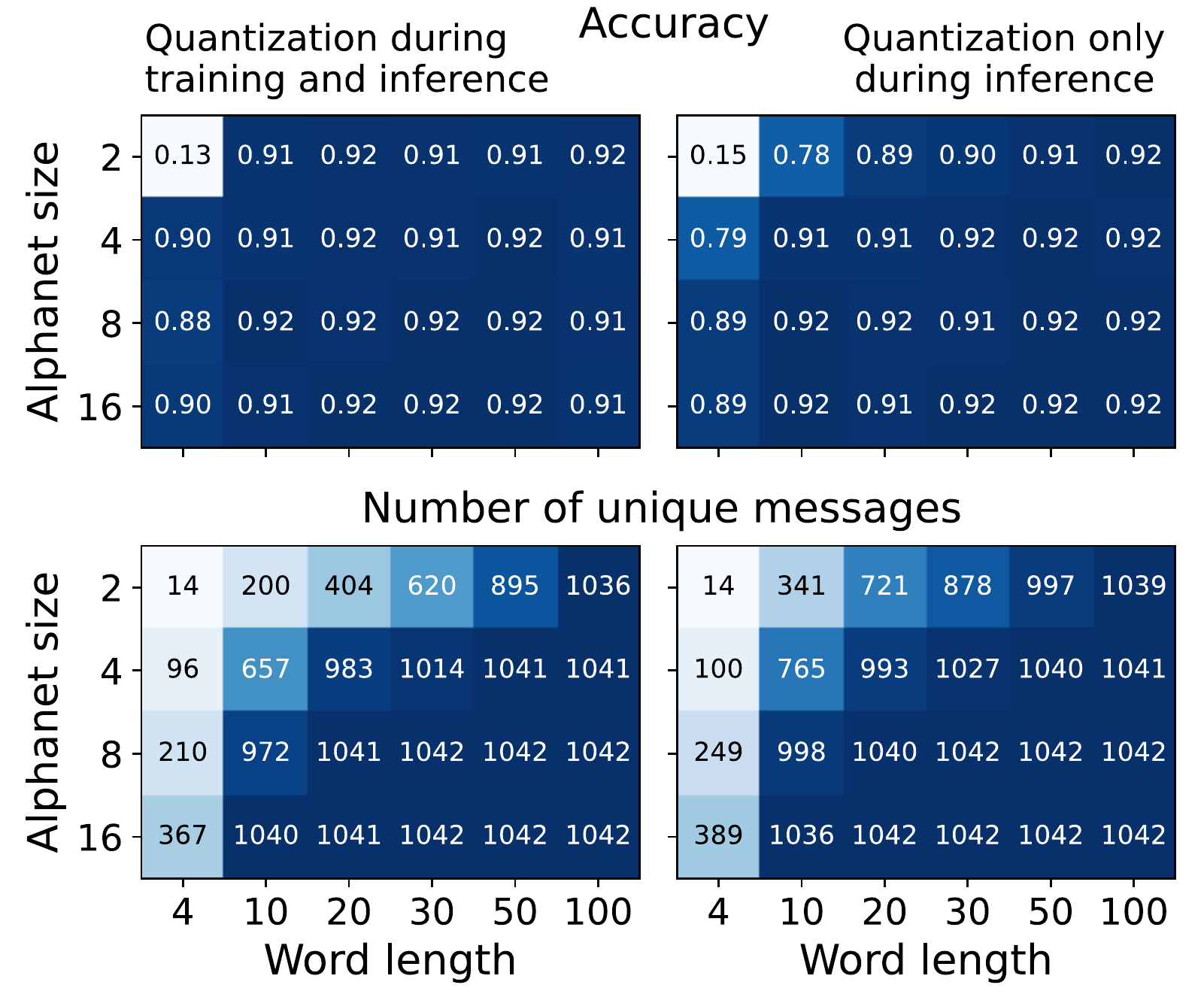}
\caption{Accuracy (top) and number of unique messages (NoUM) (bottom) as a function of alphabet size and word length for the \textbf{Sent-Cls} game with an Instant channel, comparing  quantization during both training and inference (left) or only during inference (right).
}
\label{fig:sent_cls_lang_analysis}
\end{figure}

Figures \ref{fig:image_lang_analysis}, \ref{fig:sent_ref_lang_analysis}, and \ref{fig:sent_cls_lang_analysis} provide heatmaps for the Image, Sent-Ref and Sent-Cls games, complementing the analysis from Section \ref{sec:results-lang}. 
Each figure compares accuracy results (top) and number of unique messages (bottom) for quantization during both training and inference (left) or only during inference (right).
In all games, increasing the word length significantly improves the results.
In all games, increasing the number of symbols improves the results, though less significantly.
These observations are most notifiable at the Sent-Ref game in which vertical strips are clearly seen at the upper-left accuacy heatmap of Figure~\ref{fig:sent_ref_lang_analysis}.

%In addition, in all games quantization during training degrades the performance, in line with the analysis in the main paper. 
%In the Image game (Figure~\ref{fig:image_lang_analysis}), quantization during training prevents system convergence, resulting in failure to learn. 

Interestingly, in the Sentence-Cls game (Figure~\ref{fig:sent_cls_lang_analysis}), performance reaches 91\%, only one point below the best performance achieved for this game, with a binary alphabet and word length of 10.
Correlating this with the 200 unique messages achieved for this setting, highlights the difference between the Sent-Ref and Set-Cls games.
Recall that for the Sent-Cls, 77 unique messages are enough for optimally solving the game.
%the number of unique messages is equal to the number of targets, similar to other games, while clearly for this game, it would have been sufficient to have one unique message per class, that is, 77 messages. 

\subsection{Model structure and hyper-parameters}
\label{sec:app-hparam}

Table~\ref{tbl:network-sizes} shows the sizes of neural networks used in the different games.
The Object and Image games are based on the Egg \cite{kharitonov2019egg_a20} implementation.
The Sent-Ref and Sent-Cls are based on the Egg channel architecture as well but use different agent encoders.
The Object game uses one-layer fully-connected feed-forward neural networks for the sender and receiver. 
The Image game uses one fully-connected layer, two convolutional layers, and another fully-connected layer for the sender and receiver, following the implementation in Egg. 
The Sentence games use a Distilled BERT backbone \cite{sanh2019distilbert_22}, which is frozen in the referential sentence game and fine-tuned in the classification sentence game, in order to achieve the best possible classification results. 

\begin{table}[t]
    \centering
    \resizebox{\columnwidth}{!}{
    \begin{tabular}{p{0.1cm} p{1.5cm} r r r r r r}
    \toprule 
    & & \multicolumn{2}{c}{Object} & \multicolumn{2}{c}{Image}  & \multicolumn{2}{c}{Sent-Ref/Cls}\\
\cmidrule(lr){3-4} \cmidrule(lr){5-6}   \cmidrule(lr){7-8}     

         & &  Inst & Rnn & Inst & Rnn &  Inst & Rnn\\
      \midrule 
         \multirow{8}{*}{\rotatebox[origin=b]{90}{Sender}} 
         & Channel\\
         \cmidrule(lr){2-2}
         & embedding    & --   & 1024 &  --    & 50  & --     & 768\\  
         & hidden       & W\_len & 1024 & W\_len & 20  & W\_len     & 768\\\\
         % \cmidrule(lr){3-8}
         &  Agent\\
         \cmidrule(lr){2-2}
         & embedding  & --     & --   & 50      & 50  & --    & 768\\ 
         & hidden     & W\_len & 1024 & W\_len  & 20  & W\_len & 768\\
         & kernel & --     & --   & 50      & 50  & --      & --\\ 
         \midrule
         \\
         \multirow{7}{*}{\rotatebox[origin=c]{90}{Receiver}}  
         & Channel\\
         \cmidrule(lr){2-2}
         & embedding  & --    & 1024 & --     & 50  & --    & 768\\ 
         & hidden     & W\_len & 1024 & W\_len  & 20  & W\_len & 768\\\\
         & Agent\\
         \cmidrule(lr){2-2}
         & embedding  & --     & --   & 50      & 50  & --    & 768\\ 
         & hidden     & W\_len & 1024 & W\_len  & 20  & W\_len & 768\\ 
         \bottomrule 
    \end{tabular}
    }
    \caption{Network sizes used in the different games.
    Sender's channel, sender's encoding agent, receiver channel and receiver encoding agent configuration pertained to $z_\theta$, $u_\theta$, $z_\phi$, $u_\phi$ networks as described in Figure \ref{fig:emm-comm}.
    For Instant channel, dimension of sender's channel output and receiver's channel input networks is set by the desired word length (w\_len).}
    \label{tbl:network-sizes}
\end{table}

In the experiments, we report results when varying the communication elements (alphabet size, word length, and message length).
If not mentioned specifically, we use the best tuned values. 
All hyper-parameters were tuned on the validation set of each dataset.
We found it most important to tune the Gumbel-softmax communication.
The best performing values were alphabet size of 100 (for both Instant and Recurrent channels) and message length of 6 (for Recurrent channel).
Other configurations of Gumbel-softmax which we experimented with are the temperature (best value was $t=1.0$) and the option to train with straight-through, which led to performance degradation and is thus disabled in all experiments. 

We generally found the quantized communication to be less sensitive to the the channel characteristics. The main paper reports a detailed analysis in Section \ref{sec:results-lang}. If not reported otherwise, we keep the message length in both continuous and quantized communication to be the same as in Gumbel-softmax, that is, 6 words. 

An important option is the number of negative samples (distractors) used during training, which is independent from the number of candidates used for evaluation. 
For the Object game we used all $10K$ objects, excluding the target, as negatives (distractors).
For Sent-Cls we used the maximum number of classes ($76$, excluding the target class) as negatives.
We kept this number the same for the Sent-Ref for consistency considerations.
For the Image game we use $99$ distractors, due to memory limitations.

The main training hyper-parameters we tuned were the number of epochs, the learning rate, and the batch size. 
 We trained networks for 50 epochs, which was sufficient for convergence. We allowed a patience of 50 epochs to find the best epoch, evaluated on the validation set. 
 The best learning rate was 1e-5 for Object and Sentence games, and 1e-3 for the Image game. The best batch size was 32.

\begin{table*}[h!]
\centering
% \small
%{
\begin{tabular}{l lll lrrrrrrrr}
\toprule 
& \multicolumn{3}{c}{Configuration} & & \multicolumn{8}{c}{Number of candidates} \\ 
\cmidrule(lr){2-4} \cmidrule(lr){6-13}
%\multicolumn{2}{l}{}
Comm Type & A-size  & W-len & M-len &     & 2 &  10    &  100   &  500   &  1000  &  2000  &  5000 &  10000 \\ 
\midrule 
\multirow{2}{*}{CN-RNN} & N/A     & 100   &  6    & avg &  1.000 &  1.000 &  1.000 &  1.000 &  1.000 &  1.000 &  1.000 &  1.000\\ 
       &         &       &       & std &  0.000 &  0.000 &  0.000 &  0.000 &  0.000 &  0.000 &  0.000 &  0.000\\ 
       \midrule 
\multirow{2}{*}{GS-RNN} &  10     & 10    &  6    & avg &  0.997 &  0.974 &  0.831 &  0.560 &  0.442 &  0.324 &  0.208 &  0.152\\ 
       &         &       &       & std &  0.003 &  0.021 &  0.108 &  0.223 &  0.217 &  0.189 &  0.140 &  0.110\\ 
       \midrule 
\multirow{2}{*}{QT-RNN} &  10     & 100   &  6    & avg &  1.000 &  1.000 &  1.000 &  1.000 &  1.000 &  1.000 &  1.000 &  1.000\\
       &         &       &       & std &  0.000 &  0.000 &  0.000 &  0.000 &  0.000 &  0.000 &  0.000 &  0.000\\ 
       \midrule 
%  &   &   &   &   &   &   &   &   &   &  
\multirow{2}{*}{CN-Inst} &  N/A    & 100  &  1   & avg &  1.000 &  1.000 &  1.000 &  1.000 &  1.000 &  1.000 &  0.998 &  0.998\\
       &         &      &      & std &  0.000 &  0.000 &  0.000 &  0.000 &  0.001 &  0.001 &  0.003 &  0.003\\ 
       \midrule 
\multirow{2}{*}{GS-Inst} &  100    & 100  &  1   & avg &  0.926 &  0.628 &  0.179 &  0.062 &  0.036 &  0.022 &  0.010 &  0.008\\
       &         &      &      & std &  0.004 &  0.024 &  0.003 &  0.009 &  0.002 &  0.003 &  0.005 &  0.003\\
       \midrule 
\multirow{2}{*}{QT-Inst} &  10     & 100  &  1   & avg &  1.000 &  1.000 &  1.000 &  1.000 &  1.000 &  1.000 &  1.000 &  1.000\\
       &         &      &      & std &  0.000 &  0.000 &  0.000 &  0.000 &  0.000 &  0.000 &  0.000 &  0.000\\
       \bottomrule 
 \end{tabular}
%}
%\vspace{-0.1in}
\caption{\textbf{Object} game test results for the 6 different communication types on an increasing number of candidates. Standard deviation is over three runs.}
\label{tbl:main-obj}
\end{table*}

\begin{table*}[h!]
\centering
% \small
%{
\begin{tabular}{l lll lrrrrr}
\toprule 
 & \multicolumn{3}{c}{Configuration} & & \multicolumn{5}{c}{Number of candidates} \\ 
\cmidrule(lr){2-4} \cmidrule(lr){6-10}
%\multicolumn{2}{l}{}
Comm Type & A-size  & W-len & M-len&     & 2 &  10    &  20   &  50   &  100 \\ 
\midrule 
\multirow{2}{*}{CN-RNN} & N/A     & 100   &  6      & avg & 0.993 &  0.944 &  0.893 &  0.792 &  0.675\\ 
       &         &       &         & std &  0.004 &  0.012 &  0.019 &  0.035 &  0.039\\ 
       \midrule 
\multirow{2}{*}{GS-RNN} &  50     & 50    &  6      & avg &  0.626 &  0.222 &  0.134 &  0.067 &  0.033\\ 
       &         &       &         & std &  0.209 &  0.213 &  0.150 &  0.085 &  0.043\\ 
       \midrule 
\multirow{2}{*}{QT-RNN} &  10     & 100   &  6      & avg &  0.992 &  0.948 &  0.898 &  0.786 &  0.673\\
       &         &       &         & std &  0.005 &  0.008 &  0.018 &  0.034 &  0.041\\ 
       \midrule 
%  &   &   &   &   &   &   &   &   &   &  
\multirow{2}{*}{CN-Inst} & N/A     & 100  &  1   & avg &  0.999 &  0.996 &  0.992 &  0.982 &  0.962\\
       &         &      &      & std &  0.001 &  0.003 &  0.004 &  0.008 &  0.006\\ 
       \midrule 
\multirow{2}{*}{GS-Inst} &  50     & 50   &  1   & avg &  0.951 &  0.661 &  0.472 &  0.233 &  0.132\\
       &         &      &      & std &  0.005 &  0.024 &  0.027 &  0.019 &  0.007\\
       \midrule 
\multirow{2}{*}{QT-Inst} &  10     & 100  &  1   & avg &  0.999 &  0.995 &  0.990 &  0.982 &  0.962\\
       &         &      &      & std &  0.000 &  0.001 &  0.002 &  0.006 &  0.007\\
       \bottomrule 
 \end{tabular}
%}
%\vspace{-0.1in}
\caption{\textbf{Image} game test results for the 6 different configuration on an increasing number of candidates. Standard deviation is over three runs.}
\label{tbl:main-image}
\end{table*}

\begin{table*}[h!]
\centering
% \small
%{
\begin{tabular}{l lll lrrrrrr}
\toprule 
 & \multicolumn{3}{c}{Configuration} & & \multicolumn{6}{c}{Number of candidates} \\ 
\cmidrule(lr){2-4} \cmidrule(lr){6-11}
%\multicolumn{2}{l}{}
Comm Type & A-size  & W-len & M-len &     & 2 &  8    &  16   &  32   &  64 & 77 \\ 
\midrule 
\multirow{2}{*}{CN-RNN} & N/A     & 100   &  6    & avg & 1.000 &  0.999 &  0.997 &  0.994 &  0.989 &  0.989\\ 
       &         &       &       & std &  0.000 &  0.000 &  0.003 &  0.002 &  0.001 &  0.001\\ 
       \midrule 
\multirow{2}{*}{GS-RNN} &  100    & 100    &  6    & avg &  0.814 &  0.386 &  0.226 &  0.138 &  0.078 &  0.068\\ 
       &         &       &       & std &  0.120 &  0.165 &  0.084 &  0.058 &  0.041 &  0.037\\ \midrule
\multirow{2}{*}{QT-RNN} &  10     & 100   &  6    & avg &  1.000 &  0.999 &  0.998 &  0.995 &  0.990 &  0.989\\
       &         &       &       & std &  0.000 &  0.000 &  0.002 &  0.001 &  0.002 &  0.001\\ \midrule
%  &   &   &   &   &   &   &   &   &   &  
\multirow{2}{*}{CN-Inst} &  N/A    & 100  &  1   & avg &  0.999 &  0.995 &  0.989 &  0.981 &  0.955 &  0.945\\
       &         &      &      & std &  0.001 &  0.002 &  0.002 &  0.003 &  0.009 &  0.007\\ \midrule
\multirow{2}{*}{GS-Inst} &  100    & 100  &  1   & avg &  0.673 &  0.256 &  0.139 &  0.069 &  0.039 &  0.035\\
       &         &      &      & std &  0.036 &  0.045 &  0.012 &  0.015 &  0.006 &  0.008\\\midrule
\multirow{2}{*}{QT-Inst} &  10     & 100  &  1   & avg &  0.999 &  0.996 &  0.990 &  0.981 &  0.956 &  0.945\\
       &         &      &      & std &  0.001 &  0.002 &  0.003 &  0.003 &  0.009 &  0.006\\
       \bottomrule 
 \end{tabular}
%}
%\vspace{-0.1in}
\caption{\textbf{Sentence referential (Sent-Ref)} game test results for the 6 different communication types on an increasing number of candidates. Standard deviation is over three runs.}
\label{tbl:main-sent-ref}
\end{table*}

\begin{table*}[h!]
\centering
% \small
%{
\begin{tabular}{l lll lrrrrrr}
\toprule 
 & \multicolumn{3}{c}{Configuration} & & \multicolumn{6}{c}{Number of candidates} \\ 
\cmidrule(lr){2-4} \cmidrule(lr){6-11}
%\multicolumn{2}{l}{}
Comm Type & A-size  & W-len & M-len &     & 2 &  8    &  16   &  32   &  64 & 77 \\ \midrule
\multirow{2}{*}{CN-RNN} & N/A     & 100   &  6    & avg & 0.993 &  0.972 &  0.959 &  0.943 &  0.925 &  0.920\\ 
       &         &       &       & std &  0.001 &  0.003 &  0.004 &  0.006 &  0.008 &  0.009\\ \midrule
\multirow{2}{*}{GS-RNN} &  10     & 10    &  6    & avg &  0.941 &  0.806 &  0.717 &  0.611 &  0.493 &  0.460\\ 
       &         &       &       & std &  0.023 &  0.045 &  0.045 &  0.040 &  0.032 &  0.031\\ \midrule
\multirow{2}{*}{QT-RNN} &  10     & 10    &  6    & avg &  0.993 &  0.972 &  0.959 &  0.943 &  0.925 &  0.920\\
       &         &       &       & std &  0.001 &  0.002 &  0.003 &  0.004 &  0.007 &  0.009\\ \midrule
%  &   &   &   &   &   &   &   &   &   &  
\multirow{2}{*}{CN-Inst} &  N/A    & 100  &  1   & avg &  0.994 &  0.976 &  0.964 &  0.948 &  0.927 &  0.920\\
       &         &      &      & std &  0.002 &  0.005 &  0.007 &  0.009 &  0.013 &  0.014\\ \midrule
\multirow{2}{*}{GS-Inst} &  100    & 100  &  1   & avg &  0.960 &  0.811 &  0.666 &  0.478 &  0.298 &  0.258\\
       &         &      &      & std &  0.004 &  0.014 &  0.022 &  0.033 &  0.038 &  0.038\\\midrule
\multirow{2}{*}{QT-Inst} &  10     & 100  &  1   & avg &  0.993 &  0.976 &  0.964 &  0.949 &  0.930 &  0.925\\
       &         &      &      & std &  0.002 &  0.002 &  0.004 &  0.006 &  0.008 &  0.009\\
       \bottomrule 
 \end{tabular}
%}
%\vspace{-0.1in}
\caption{\textbf{Sentence classification (Sent-Cls)} game test results for the 6 different communication types on an increasing number of candidates. Standard deviation is over three runs.}
\label{tbl:main-set-cls}
\end{table*}

\end{document}